\documentclass[11pt]{article}
\PassOptionsToPackage{sort&compress}{natbib}
\usepackage[final]{acl_arxiv}
\usepackage{times}
\usepackage{latexsym}
\usepackage[T1]{fontenc}
\usepackage[utf8]{inputenc}
\usepackage{microtype}
\usepackage{inconsolata}
\usepackage{graphicx}
\usepackage{algorithm}
\usepackage{float}
\usepackage{algpseudocode}
\usepackage{multirow}
\usepackage{makecell}
\usepackage{amsmath}
\usepackage{url}
\usepackage{booktabs}
\usepackage{amsfonts}
\usepackage{nicefrac}
\usepackage{xcolor}

\title{Drive As You Like: Multi-Head Diffusion with Reinforcement Learning for Personalized Driving}

\author{
  \textbf{Ding Fan}\textsuperscript{1} \quad
  \textbf{Xuewen Luo}\textsuperscript{2} \quad
  \textbf{Hwa Hui Tew}\textsuperscript{1} \\
  \textbf{Susilawati}\textsuperscript{1} \quad
  \textbf{Vishnu Monn Baskaran}\textsuperscript{1,\ensuremath{\dagger}} \quad
  \textbf{Fucai Ke}\textsuperscript{3,\ensuremath{\dagger}} \\
  \textsuperscript{1}Monash University, Malaysia \\
  \textsuperscript{2}University of Utah, USA \\
  \textsuperscript{3}Monash University, Australia \\[0.5ex]
  \textsuperscript{\ensuremath{\dagger}}These authors contributed equally to the supervision of this work.
}

\begin{document}
\setcitestyle{numbers,square,comma}
\maketitle

% \maketitle

\begin{abstract}
Personalized autonomous driving requires understanding the driving needs implicit in user requests.  This is challenging when driving demonstrations lack style annotations and behavioral adaptation risks degrading planning quality. We propose M-Diffusion Planner with a dual-decoupling design: separating intent understanding from trajectory planning, and general planning capability learning from strategy alignment. A large language model infers driving needs and selects a strategy, which is executed by the corresponding head of a shared diffusion planner. Following imitation pretraining, strategy-specific rewards specialize individual heads without style annotations. We innovatively adapt Group Relative Policy Optimization using surrogate policy-ratio estimation and constrained head-specific updates to mitigate degradation of pretrained planning capabilities. Open-loop and closed-loop evaluations on nuPlan demonstrate distinct strategy-aligned behaviors and competitive planning performance. Further experiments assess intent-to-strategy mapping and collision risk, and analyze the effects of reward design and optimization on strategy alignment and planning quality.
\end{abstract}

\section{Introduction}

Modern autonomous driving systems have achieved substantial progress in core modules such as perception~\cite{zhao2024bev,liu2025vision}, localization~\cite{qin2021light,zheng2023simultaneous}, and control~\cite{vu2021model,hang2021towards,ke2025explain}, enabling stable operation in diverse and complex traffic environments. However, existing planners typically overlook diverse and context-dependent user preferences~\cite{zheng2024preliminary}. While safety remains the primary constraint, real-world driving often involves nuanced trade-offs between comfort and efficiency. For example, transporting pregnant passengers or fragile goods requires smoother and more stable driving, whereas time-sensitive travel may prioritize efficiency while maintaining safety guarantees.

On the planning side, personalized driving faces two major bottlenecks: a lack of explicit style annotations and the difficulty of generating distinct driving behaviors aligned with user preferences. 
Specifically, many existing autonomous driving datasets lack explicit style annotations~\cite{hao2026styledrive}, making it difficult to learn a direct mapping between user preferences and driving behaviors. 
Existing methods introduce behavioral control through guidance during trajectory generation. For example, Diffusion-ES uses LLM-generated rewards to guide driving~\cite{yang2024diffusion}, Diffusion Planner applies inference-time guidance to adjust speed and comfort~\cite{zheng2025diffusion}, and CTG++ translates language instructions into guidance objectives for traffic simulation~\cite{zhong2023language}. 
While these methods enable control over generated trajectories, learning distinct driving strategies aligned with user preferences remains challenging. In particular, the multimodality of diffusion models does not automatically associate different trajectory modes with interpretable driving styles. The challenge is therefore to learn this correspondence without explicit style annotations, enabling users to select driving strategies that reflect their preferences.

Reinforcement learning (RL) offers a viable way to establish a correspondence between user preferences and driving behaviors through reward-based optimization~\cite{surmann2025multi,ke2024hydra}. By encoding driving preferences as reward objectives, planners can learn preference-aligned policies from reward feedback, reducing their reliance on explicit style annotations. 
However, learning high-quality driving behaviors directly through RL is often sample-intensive and requires careful reward design~\cite{huang2022efficient}. Moreover, optimizing a policy for a single fixed preference objective does not directly support selection among different driving strategies. These limitations motivate building on pretrained planning capabilities and aligning them with multiple preference-specific objectives. Applying such alignment to diffusion planners introduces two further technical challenges: the intractable marginal likelihood of generated trajectories complicates probability-ratio-based policy updates~\cite{black2024training}, while preference-driven fine-tuning can degrade pretrained planning capabilities~\cite{nakamoto2023cal}.

\begin{figure*}[t]
    \centering
    \includegraphics[width=\textwidth]{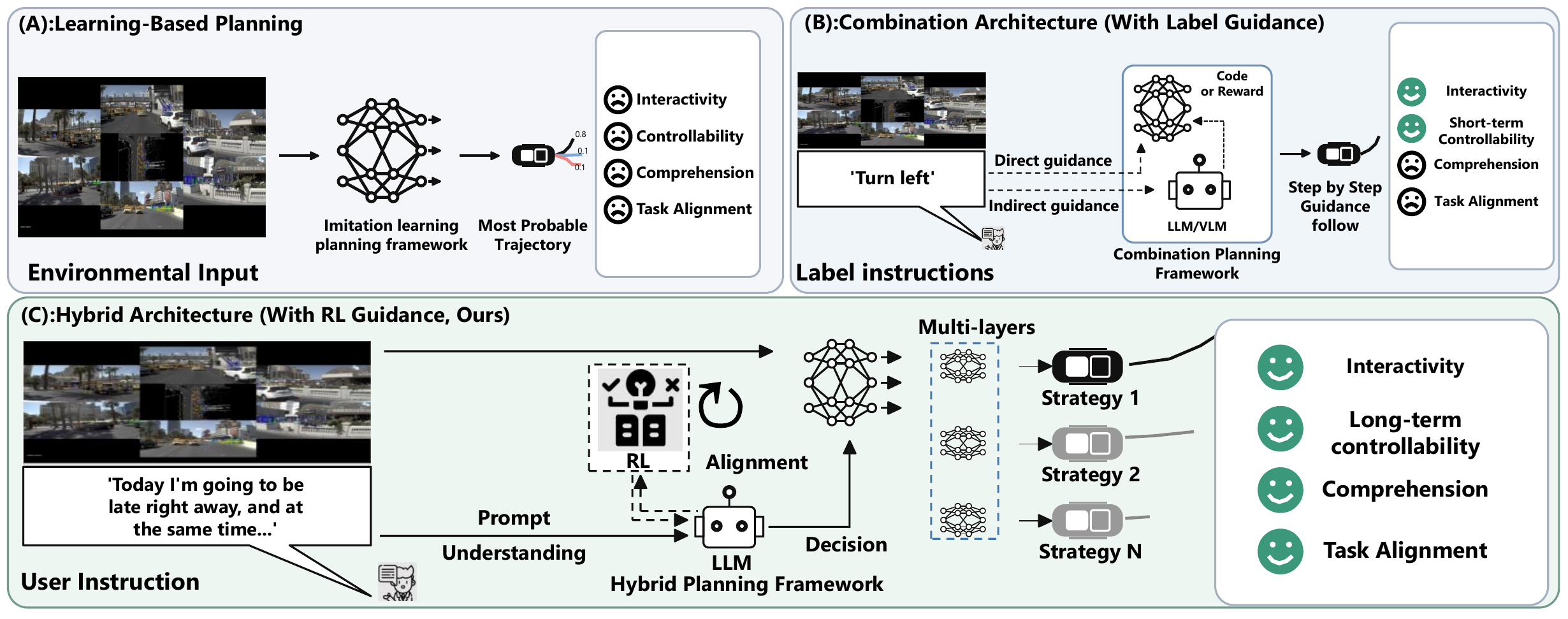}
    \caption{
    Comparison of autonomous driving planning paradigms.
    (A) Imitation learning directly maps inputs to a single trajectory, lacking interactivity and controllability.
    (B) Label-guided methods introduce external instructions for short-term control, but remain limited in understanding and long-term alignment.
    (C) Our hybrid framework combines LLM-based intent understanding with RL-guided planning, enabling multi-strategy trajectory generation with improved interactivity and alignment.
    }
    \label{fig:planning_paradigms}
\end{figure*}

Based on these observations, we view personalized driving as a complex task involving nested, sequentially connected subtasks whose objectives are difficult to jointly accommodate within a single model. We therefore propose M-Diffusion Planner with a dual-decoupling design, as illustrated in Fig.~\ref{fig:planning_paradigms}: first, separating user-intent interpretation from diffusion-based trajectory planning; and second, separating foundational planning capability learning from strategy alignment within the planner. This design decomposes the tightly coupled task into interconnected subtasks with distinct objectives, allowing each stage to pursue its own goal while coordinating their outputs to promote consistent overall driving behavior.

Following this decoupling, M-Diffusion Planner assigns the user-intent understanding task to an LLM, which bridges users and the planner by interpreting requests in context and mapping them to discrete driving strategies. The subsequent planning task is further decomposed into foundational capability learning and strategy alignment. For the first subtask, we pretrain a base diffusion planner on expert trajectories through imitation learning and initialize multiple strategy-specific output heads from the same pretrained model. 
% For the second, we align each head with its target driving preference through RL post-training using strategy-specific rewards. This optimization faces two challenges: the marginal likelihood of diffusion-generated trajectories is generally intractable, complicating probability-ratio-based updates~\cite{black2024training}, and RL fine-tuning can degrade pretrained capabilities~\cite{nakamoto2023cal}. 
For RL post-training, we adapt GRPO using a Monte Carlo-fitted Gaussian surrogate for policy-ratio estimation. To mitigate planning-performance degradation, we update only the selected strategy head while freezing the shared backbone and other heads, and regularize optimization through surrogate-ratio clipping and Gaussian KL penalties.

After addressing the individual subtasks, we integrate the corresponding modules through clearly defined interfaces to form a unified framework for the complex task of personalized autonomous driving. Our main contributions are summarized as follows:

\begin{itemize}
    \item We propose M-Diffusion Planner, a two-stage framework that decouples general planning capability learning from preference alignment and combines strategy-specific output heads with user intent understanding to enable controllable driving strategies.

    \item Building on GRPO, we design trajectory likelihood approximation and constrained policy update mechanisms for diffusion planner post-training, enabling preference alignment while mitigating degradation of general planning capabilities.

    \item We evaluate M-Diffusion Planner on the nuPlan benchmark to assess preference alignment across driving strategies and trajectory planning quality. The results demonstrate that our method produces distinct preference-aligned driving behaviors while maintaining competitive overall planning performance.
\end{itemize}

\section{Related Work}

\subsection{Motion planner}
Early motion planning methods in autonomous driving were primarily designed around handcrafted heuristics and control rules~\cite{xiao2021rule,ding2024energy}, which ensured safety and feasibility in structured environments such as highways or lane-following scenarios. These methods struggled in dynamic urban contexts due to limited adaptability and scalability. With the rise of data-driven approaches, learning-based planners began leveraging imitation learning to predict future trajectories directly from human driving data, using neural architectures such as Convolutional Neural Networks (CNNs)~\cite{hoshino2022motion}, Recurrent Neural Networks (RNNs)~\cite{jeong2020surround}, and hybrid CNN–LSTM frameworks~\cite{xiao2021rule}. More recently, Transformer-based planners equipped with self-attention mechanisms~\cite{lu2025transformer,ke2025dwim,huang2023gameformer,li2021multi,mebrahtu2023transformer} have demonstrated strong capability in modeling long-range dependencies and handling complex, multi-agent interactions. Despite these advances, current rule-based or learning-driven planners face challenges in real-time adaptability, controllability, and effective human–planner interaction in dynamic driving environments.

\subsection{Human-Interactive Planning}
Recent advances explore controllable trajectory generation using diffusion models. CTG++~\cite{zhong2023language} uses an LLM to translate natural-language requests into loss functions that guide a scene-level diffusion model for controllable traffic simulation. Diffusion-ES\cite{yang2024diffusion} employs a gradient-free optimization strategy combined with trajectory denoising, enabling black-box reward optimization and achieving state-of-the-art results in autonomous driving and zero-shot instruction following. Furthermore, several studies ~\cite{zheng2025diffusion,lin2025causal} have explored classifier-free guidance methods, incorporating environmental contexts directly into the diffusion process to enhance adaptability and control during trajectory generation. Diffusion-Planner\cite{zheng2025diffusion} adopts a classifier-free guidance strategy to align trajectory generation with human preferences. However, although these methods can reflect human intentions to a certain extent, their interaction mechanisms are still primarily limited to short-term guidance or single-step conditioning~\cite{ke2026view2space, ma2025position}. They lack the ability to maintain consistent long-term interaction and continuously adapt planning behaviors according to evolving human intentions over extended driving horizons.

\subsection{Reinforcement learning Post-training}
Reinforcement learning (RL)~\cite{wiering2012reinforcement} has been widely applied to decision-making, planning, and control in autonomous driving ~\cite{brunke2022safe}. RL methods are generally divided into two categories. The first is value-based methods, which select actions by maximizing a learned value function, such as Q-learning and deep Q-learning. The second is policy-based methods, which directly optimize a parameterized policy, with representative algorithms including PPO~\cite{schulman2017proximal}. Although policy-based approaches can train probabilistic output models, they typically incur high computational complexity and substantial training cost. GRPO~\cite{guo2025deepseek} mitigates this issue by removing the critic component in PPO and replacing value estimation with explicit reward signals, thereby simplifying policy optimization.

\section{Method}
We first define the strategy driving task and clarify the input–output representations in Section~\ref{sec:task_definition}. In Section~\ref{sec:model_architecture}, we introduce the overall architecture of our model, which is built upon a multi output layer diffusion planner based on diffusion transformer (DiT)~\cite{zheng2025diffusion}. In Sections~\ref{sec:pretraining} and ~\ref{sec:posttraining}, we present how the model is pre-trained and post-trained under different strategies, and describe the primary cost functions.

\begin{figure*}[t]
  \centering
  \includegraphics[width=\textwidth]{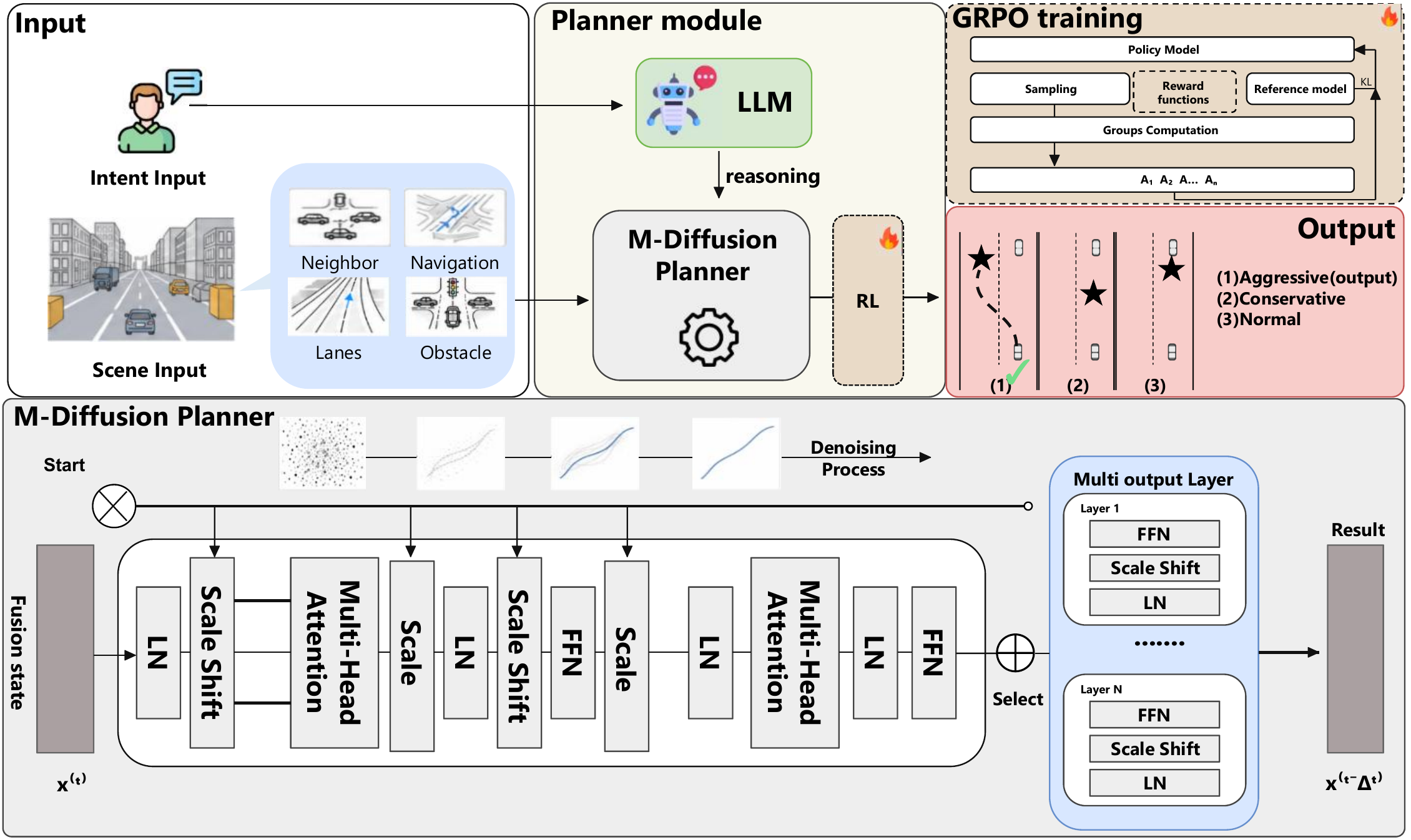}
  \caption{Model architecture of M-Diffusion Planner.}
  \label{fig:overview}
\end{figure*}

\subsection{Task definition}\label{sec:task_definition}
In autonomous driving, planning should consider not only spatiotemporal information but also human needs and preferences. We treat these preferences as strategy-level intentions, reflected in different driving styles. Each strategy corresponds to a specific user command or high-level behavioral goal. Guiding the planner through strategy-level inputs, rather than low-level controls, improves generalization across scenarios and enables more flexible, user-aligned driving behavior.

To learn the underlying distribution of feasible trajectories, we adopt a diffusion-based denoising objective. Let $x^{(0)}$ denote the ground-truth trajectory and $x^{(t)}$ its noisy counterpart at diffusion step $t$. The denoising network takes the scene context $C_1$ as input. The LLM maps the user's request to a discrete driving strategy $C_2$, which selects the corresponding output head $h(C_2)$. The selected head predicts the clean trajectory as $\mu_{\theta,h(C_2)}(x^{(t)},t,C_1)$. The denoising objective is formulated as:

\begin{equation}
\resizebox{0.90\columnwidth}{!}{$\displaystyle
\begin{aligned}
\mathcal{L}_\theta = 
\mathbb{E}_{x^{(0)},\, t,\, x^{(t)}} \Big[
\big\|
\mu_{\theta,h(C_2)}(x^{(t)}, t, C_1)
- x^{(0)}
\big\|^2
\Big], \\
t \sim \mathcal{U}(0,1), \qquad
x^{(t)} \sim q_{t}(x^{(t)}|x^{(0)}).
\end{aligned}
$}
\label{eq:training_loss}
\end{equation}

Based on the model prediction of the clean trajectory 
$\hat{x}_0 = \mu_{\theta,h(C_2)}(x^{(t)}, t, C_1)$, the corresponding score 
function under the denoising process is given by
\begin{equation}
\label{eq:score_function}
s_\theta =
\frac{\alpha_t\, 
\mu_{\theta,h(C_2)}(x^{(t)}, t, C_1)
- x^{(t)}
}{\sigma_t^{2}}.
\end{equation}
where $\alpha_t$ and $\sigma_t$ are the signal and noise coefficients of the forward diffusion process. Using this estimated score, the reverse diffusion process can be expressed in 
discrete form as:
\begin{equation}
\resizebox{0.90\columnwidth}{!}{$\displaystyle
\begin{aligned}
x^{(t-\Delta t)}
&=
\frac{\alpha_{t-\Delta t}}{\alpha_t}\, x^{(t)}
+
\left(
\frac{\alpha_{t-\Delta t}\,\sigma_t^{2}}{\alpha_t}
-
\sigma_{t-\Delta t}\,\sigma_t
\right)
s_\theta\
\\[4pt]
&\quad
+\,\sigma_{t-\Delta t}\, z,
\qquad z\sim\mathcal{N}(0,I).
\end{aligned}
$}
\label{eq:reverse_step_score}
\end{equation}
where the above formulation corresponds to the discrete-time reverse 
transition used to gradually refine the noisy sample towards a feasible 
clean trajectory.

\subsection{Model Architecture}\label{sec:model_architecture}
Figure~\ref{fig:overview} presents the architecture and training workflow of M-Diffusion Planner. The model consists of a scene encoder based on MLP-Mixer and Transformer blocks, and a DiT-based decoder with a shared backbone and strategy-specific output heads. The scene encoder provides contextual representations, while the selected output head predicts clean trajectories corresponding to the requested driving strategy.

The planner combines multiple sources of scene information. Noisy future trajectories are concatenated with the current states of the ego vehicle and neighboring agents, enabling self-attention to model interactions among traffic participants. Historical neighboring-agent and lane features are encoded through token and channel mixing with residual connections. Static-object features are encoded by MLP and Transformer blocks and incorporated through multi-head cross-attention. Navigation features are encoded using an MLP-Mixer and combined with the diffusion timestep embedding to guide trajectory generation along the intended route.

\subsection{Multi-Strategy Head Pre-training}\label{sec:pretraining}

M-Diffusion Planner is a DiT-based~\cite{peebles2023scalable, zheng2025diffusion} multi-output-layer architecture, designed to accommodate diverse user requirements by generating motion plans aligned with specified driving strategies. A limited set of strategy structures is a reasonable design choice, as common driving preferences can be represented by a finite set of high-level strategies~\cite{hao2026styledrive}.

To reduce training cost and improve efficiency, we adopt a weight-sharing initialization scheme. Specifically, we first train a single-output diffusion planner with an identical backbone architecture, and subsequently transfer its learned weights to initialize the multi-head M-Diffusion Planner. This enables efficient knowledge reuse while avoiding the high computational overhead of training a multi-head model from scratch. During inference, the model conditions on both scene inputs and strategy specifications to generate trajectories, outputting predicted $(X, Y)$ positions and heading corresponding to different driving strategies.

\subsection{GRPO-Based Post-Training}\label{sec:posttraining}

After completing pre-training, we ensure that each strategy head is capable of generating high-quality trajectories. We then freeze all model parameters except for the output head currently being trained, including shared components and non-target strategy heads, and fine-tune the selected head using GRPO. During this stage, a strategy-specific reward function is introduced to evaluate trajectories sampled from the diffusion distribution, enabling targeted policy optimization for each strategy. In this process, the model is post-trained using predefined reward functions rather than relying on collected data. This approach avoids the complex procedures involved in gathering and categorizing driving data, while ensuring that the fine-tuning remains closely aligned with the intended driving strategies.

During GRPO-based training, only the output layer of the selected strategy head is updated, while all other strategy heads and shared model parameters remain frozen. 
Given an input batch, the encoder produces a latent representation $z$. 
Conditioned on $z$ and the selected strategy $s$, the decoder samples $G$ candidate trajectories 
$\{\hat{\tau}_i\}_{i=1}^{G}$, where $G$ denotes the number of Monte Carlo samples. Each trajectory is evaluated by a reward function $R(\hat{\tau}_i)$, yielding a scalar reward $r_i$, where $r_i = R(\hat{\tau}_i)$.

For strategy $s$, the reward is defined as
\begin{equation}
\begin{aligned}
R_s(\tau)
={}& w_v^{(s)}\phi_v(\bar{v})
  + w_a^{(s)}\phi_a(\bar{a}) \\
 &+ w_j^{(s)}\phi_j(\bar{j})
  + w_d^{(s)}\phi_d(\bar{d}),
\end{aligned}
\label{eq:strategy_reward}
\end{equation}
Here, $\bar{v}$, $\bar{a}$, $\bar{j}$, and $\bar{d}$ are the
trajectory's mean velocity, acceleration magnitude, jerk magnitude,
and following distance. $\phi_k$ normalizes component $k$, and
$w_k^{(s)}$ is its strategy-specific weight.

% For example, under an aggressive strategy, higher weights are assigned to velocity and acceleration to encourage faster and more dynamic trajectories. In contrast, a comfort-oriented strategy places lower weights on jerk to promote smoother driving behavior.

The sampled rewards are standardized into trajectory-level advantages:
\begin{equation}
\begin{aligned}
A_i &= \frac{r_i - \mu_r}{\sigma_r + \epsilon}, \\
\mu_r &= \mathrm{mean}(r_{1:G}), \quad
\sigma_r = \mathrm{std}(r_{1:G}).
\end{aligned}
\label{eq:adv_norm}
\end{equation}
\noindent
where $A_i$ denotes the normalized advantage of the $i$-th sampled trajectory, 
$\mu_r$ and $\sigma_r$ are the sample mean and standard deviation of rewards 
$\{r_i\}_{i=1}^{G}$, and $\epsilon$ is a small constant for numerical stability.

Multiple sampled trajectories within the same group jointly contribute to the policy update. 
The overall GRPO objective is formulated as:
\begin{equation}
\begin{aligned}
\mathcal{J}_{\text{GRPO}}(\theta)
&=
\mathbb{E}
\left[
\frac{1}{G}
\sum_{i=1}^{G}
\mathcal{L}_i^{\text{GRPO}}
\right], \\
&\quad q \sim p(Q),\quad
\hat{\tau}_{1:G}
\sim \pi_{\theta_{\text{old}}}(O \mid q).
\end{aligned}
\label{eq:grpo1}
\end{equation}
Here, we define \(p(Q)\) as the distribution over task queries used during policy optimization. 
The query \(q=\{C_1, C_2\}\) denotes the task-specific condition consistent with Eq.~(1). 
For each query \(q\), we sample a group of \(G\) denoised trajectories 
\(\{\hat{\tau}_i\}_{i=1}^{G} \sim \pi_{\theta_{\text{old}}}(O \mid q)\), 
where each \(\hat{\tau}_i\) corresponds to the final clean trajectory produced by the diffusion sampler.
The policy \(\pi_{\theta_{\text{old}}}\) is a frozen copy of the current policy from the previous epoch, used solely for trajectory sampling to ensure a stable reference during optimization.

The group-wise objective $\mathcal{L}_{i}^{\text{GRPO}}$ is defined at the \emph{trajectory level} as:

\begin{equation}
\begin{aligned}
\mathcal{L}_{i}^{\text{GRPO}}
={}& \min \Big(
\rho_i A_i,\,
\text{clip}(\rho_i,1-\varepsilon,1+\varepsilon)A_i
\Big) \\
&- \beta\,\mathbb{D}_{\text{KL}}\!\left(
\pi_{\theta}\ \|\ \pi_{\text{ref}}
\right).
\end{aligned}
\label{eq:grpo2}
\end{equation}

 where \texttt{clip} denotes the clipping operator that constrains the policy update and prevents overly aggressive changes to the policy. 
The parameter $\varepsilon$ is a small clipping constant that controls the maximum allowable deviation between the updated policy and the old one, while the second term regularizes the updated policy $\pi_{\theta}$ toward a reference policy $\pi_{\text{ref}}$ via a KL Divergence penalty weighted by $\beta$. 
With the trajectory-level policy ratio
\begin{equation}
\tilde{\rho}_i =
\frac{\tilde{p}_{\theta}(\hat{\tau}_i \mid q)}
{\tilde{p}_{\theta_{\mathrm{old}}}(\hat{\tau}_i \mid q)},
\qquad
\hat{\tau}_i \sim \pi_{\theta_{\mathrm{old}}}.
\label{eq:grpo3}
\end{equation}

where the old policy $\pi_{\theta_{\text{old}}}$ is initialized from the pretrained baseline model 
and remains fixed within each training epoch, 
whereas the current policy $\pi_{\theta}$ is iteratively updated by gradient ascent on $\mathcal{J}_{\text{GRPO}}(\theta)$. 
After each epoch, $\pi_{\theta_{\text{old}}}$ is synchronized with the updated $\pi_{\theta}$ to establish a new trust region for the next iteration.
Since the exact marginal likelihood of a diffusion model is generally intractable, and 
$\rho_i$ serves only to quantify the \textbf{relative deviation} between policies rather than their 
absolute likelihoods, we estimate the trajectory likelihood by fitting a Gaussian distribution 
over $S$ Monte Carlo trajectory samples:
\begin{equation}
\tilde{p}_{\theta}(\tau \mid q)
=
\mathcal{N}\left(
\tau \mid \mu_\theta(q), \Sigma_\theta(q)
\right),
\label{eq:mc_gaussian}
\end{equation}

where $\mu_\theta(q)$ and $\Sigma_\theta(q)$ are the Monte Carlo-estimated trajectory mean and diagonal covariance. For each condition $q$, we fit separate Gaussian surrogates to $S$ trajectories sampled from the current and old policies. Both surrogates are evaluated at the same candidate trajectory $\hat{\tau}_i$ sampled from the old policy to estimate $\tilde{\rho}_i$.

\subsection{LLM Prompt Engineering}

We design tailored prompts to enable a pretrained LLM to map explicit user commands and implicit task requirements to one of four predefined driving strategies: base, aggressive, conservative, and comfortable. The LLM is invoked at the beginning of each task and again when the user's intent changes. Its output is encoded as a discrete strategy ID and deterministically mapped to the corresponding output head through a fixed rule. The selected head remains active for subsequent planning steps until the strategy is updated, keeping LLM inference outside the diffusion denoising loop. On a benchmark of 1,000 human-intent pairs with ground-truth strategy labels, Gemma 3 12B~\cite{team2025gemma}, GPT-4o~\cite{hurst2024gpt}, and Laguna S 2.1~\cite{poolside2026laguna} achieve classification accuracies of 97.1\%, 96.9\%, and 97.6\%, respectively, using the same prompt and label space (Appendix Table~\ref{tab:intent_accuracy}).

\section{Experiments}

We evaluate M-diffusion planner ability to generate distinct driving behaviors aligned with user preferences while preserving general planning capabilities. Our evaluation covers strategy alignment, closed-loop planning performance, intent understanding, and collision risk. We further investigate how reward design and RL optimization influence the balance between behavioral specialization and planning quality. Additional experimental details and results are provided in the Appendix.

\subsection{Open-loop Strategic alignment Evaluation}
In the open-loop evaluation, we utilize the outputs from different decoder heads to assess the distinctiveness among strategies, with a particular focus on velocity, acceleration, and jerk. To ensure a comprehensive evaluation, we conduct open-loop testing across $2,000$ recorded scenarios. This large-scale setup allows us to systematically examine the behavioral variations across strategies and mitigates the influence of randomness or isolated cases.

\textbf{Experimental Results.} The open-loop experimental results are presented in Table ~\ref{tab:open_loop_eval}. We evaluate the outputs under different policy heads. Specifically, under the aggressive driving strategy, the generated trajectories exhibit higher speeds, accompanied by increased acceleration. In contrast, the conservative strategy leads to lower speeds and smoother vehicle dynamics. Notably, the comfort-oriented strategy produces trajectories with the lowest jerk, which is consistent with the objective of enhancing ride smoothness and stability.
\begin{table*}[ht]
\centering
\small
\renewcommand{\arraystretch}{1.1}
\caption{Open-loop evaluation on 2,000 scenarios comparing different M-Diffusion planners in terms of motion and velocity distributions. We analyze velocity, acceleration, and jerk under different fine-tuning strategies, and group trajectories into low (0–10 m/s), medium (10–20 m/s), and high ($>20$ m/s) speed ranges for fair behavioral comparison.}
\label{tab:open_loop_eval}
\resizebox{\textwidth}{!}{  
\begin{tabular}{lcccccc}
\toprule
\textbf{Method} & \textbf{Velocity (m/s)} & \textbf{Acceleration (m/s\textsuperscript{2})} & \textbf{Jerk (m/s\textsuperscript{3})} & \textbf{Low Speed (\%)} & \textbf{Mid Speed (\%)} & \textbf{High Speed (\%)} \\
\midrule
M-Diffusion Planner (Base) & 10.59 & 1.97 & 2.66 & 49.61 & 33.85 & 16.54 \\
M-Diffusion Planner (Aggressive) & 12.50 & 2.31 & 2.43 & 43.49 & 29.95 & 26.56 \\
M-Diffusion Planner (Conservative) &  9.57 & 1.80 & 2.58 & 56.51 & 35.81 & 7.68 \\
M-Diffusion Planner (Comfortable) & 11.03 & 1.72 & 1.85 & 47.1& 35.2 & 17.7 \\
\bottomrule
\end{tabular}
}
\end{table*}

These results demonstrate that the proposed model is capable of generating trajectories with distinct characteristics conditioned on the specified driving strategies, validating its effectiveness in strategy-aware planning and providing a foundation for enabling efficient human--machine interaction.
\begin{figure*}[!t]
    \centering
    \includegraphics[width=\textwidth]{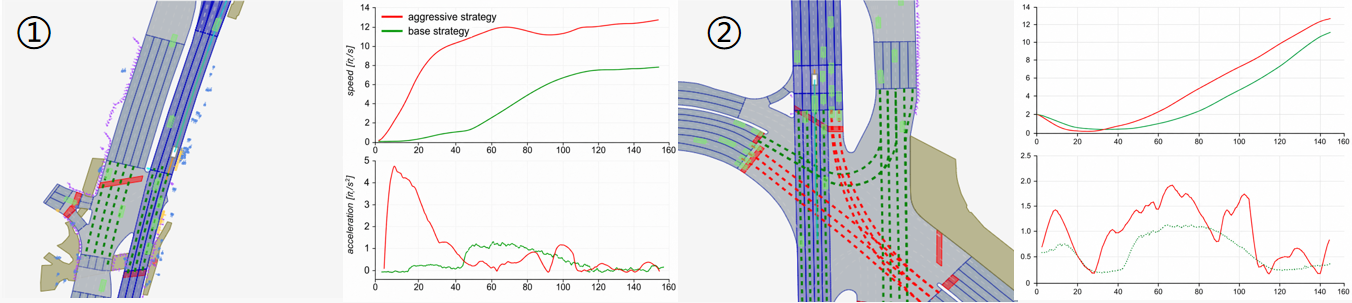}
    \caption{Comparison between the aggressive and base strategies
    in two representative driving scenarios.
    The upper plots show velocity profiles, and the lower plots
    show acceleration magnitudes, with time steps on the horizontal
    axis. Additional cases are provided in the Appendix.}
    \label{fig:case_study}
\end{figure*}

\subsection{Closed-Loop Planning Performance}

We evaluate our planner on closed-loop nuPlan simulation tasks under both non-reactive and reactive settings. Compared with the evaluated baselines, which are reported without user-intent alignment, our framework provides selectable, preference-aligned driving strategies while retaining competitive planning performance, with some strategy configurations approaching state-of-the-art scores.

\begin{table*}[ht!]
\centering
\caption{Closed-loop evaluation on nuPlan. Bold: column best; ``/'' denotes results unavailable in the published baseline comparisons; *: pre-searched reference lines or additional proposals used as inputs.}
\label{tab:Closed_loop_eval}
\resizebox{\textwidth}{!}{
\begin{tabular}{lcccccc}
\toprule
\textbf{Method}
& \textbf{Val14 (NR)}
& \textbf{Val14 (R)}
& \textbf{Test14-hard (NR)}
& \textbf{Test14-hard (R)}
& \textbf{Test14 (NR)}
& \textbf{Test14 (R)} \\
\midrule
\multicolumn{7}{c}{\textit{Learning-based Methods}} \\
\midrule
PDM-Open*~\cite{dauner2023parting}
& 53.53 & 54.24 & 33.51 & 35.83 & 52.81 & 57.23 \\
UrbanDriver~\cite{scheel2022urban}
& 68.57 & 64.11 & 50.40 & 49.95 & 51.83 & 67.15 \\
GameFormer~\cite{huang2023gameformer} w/o refine.
& 13.32 & 8.69 & 7.08 & 6.69 & 11.36 & 9.31 \\
PlanTF~\cite{cheng2024rethinking}
& 84.72 & 76.95 & 69.70 & 61.61 & 85.62 & 79.58 \\
PLUTO~\cite{cheng2024pluto} w/o refine.*
& 88.89 & 78.11 & 70.03 & 59.74 & \textbf{89.90} & 78.62 \\
Diffusion-es w/o LLM~\cite{yang2024diffusion}
& 50.00 & / & / & / & / & / \\
STR2-CPKS-800M~\cite{sun2025generalizing} w/o refine.*
& 65.16 & / & 52.57 & / & 68.74 & / \\
Diffusion Planner~\cite{zheng2025diffusion}
& \textbf{89.87} & 82.80 & \textbf{75.99}
& \textbf{69.22} & 89.19 & \textbf{82.93} \\
\midrule
\multicolumn{7}{c}{\textit{M-Diffusion Planner (Ours)}} \\
\midrule
M-Diffusion Planner (Base)
& 88.73 & 81.65 & 74.87 & 68.71 & 87.21 & 81.33 \\
M-Diffusion Planner (Aggressive)
& 82.63 & 75.11 & 68.21 & 63.83 & 74.26 & 73.24 \\
M-Diffusion Planner (Conservative)
& 85.51 & 78.69 & 69.72 & 65.71 & 76.02 & 75.21 \\
M-Diffusion Planner (Comfortable)
& 88.72 & \textbf{83.31} & 70.35 & 67.83 & 81.25 & 77.27 \\
\bottomrule
\end{tabular}
}
\end{table*}

\textbf{Baselines.} The baseline methods are categorized into three types based on their training paradigms\cite{dauner2023parting}: rule-based, learning-based, and hybrid approaches. To ensure a fair and comprehensive comparison, we utilize the trajectories generated by each planner and apply physics-consistent post-processing using the built-in NuPlan simulator. This process converts raw trajectories into complete state sequences, including timestamps, velocity, acceleration, and orientation, making them suitable for downstream control or evaluation.

\textbf{Experimental Results.} Table~\ref{tab:Closed_loop_eval} presents the evaluation results on the nuPlan simulation benchmarks. The base planner initially achieves performance close to the state-of-the-art level. After applying different policy-specific fine-tuning strategies, all models consistently maintain high NuPlan scores. These results demonstrate that the GRPO-based fine-tuning framework effectively preserves the planner’s core capabilities while enabling behavior adaptation aligned with different driving preferences. Although the NuPlan scores exhibit a slight decrease compared with the base model, this behavior is expected because GRPO primarily optimizes strategy-specific rewards rather than directly maximizing the NuPlan benchmark metrics. Consequently, a natural trade-off emerges between specialized policy behaviors and overall benchmark performance.

GRPO-based strategy alignment involves a trade-off between personalization and planning performance. Directly post-training the planner with REINFORCE~\cite{huang2022efficient} using the same reward functions and weights leads to substantial degradation of pretrained planning capabilities, with the nuPlan score dropping to 61.27 after three epochs and below 20 after ten epochs (Table~\ref{tab:reinforce_results} in the Appendix).

\subsection{User Intent Understanding}
To illustrate strategy selection through human--machine interaction, we consider a commuting scenario in which the user provides the high-level instruction, ``I am running late for work.'' The LLM interprets this request as a preference for travel efficiency and selects the \textit{aggressive strategy}. We evaluate the planner under this strategy across four representative closed-loop scenarios: (1) simple straight driving, (2) complex straight driving, (3) car-following turning, and (4) low-speed waiting at traffic lights. Figure~\ref{fig:case_study} presents two representative cases, with the remaining cases provided in the Appendix. 

In case (1), the aggressive strategy accelerates earlier and reaches a higher speed than the base strategy, with a pronounced initial acceleration peak followed by lower acceleration magnitudes. In case (2), both strategies initially reduce speed, after which the aggressive strategy increases speed more rapidly and maintains a higher velocity profile. Its acceleration profile also exhibits larger peaks and greater fluctuations than the base strategy. These cases illustrate how the selected strategy changes the planner's speed and acceleration behavior in line with the efficiency-oriented request.

\subsection{Effect of Reward Design}
Explicit rewards induce stronger speed-based alignment with the conservative strategy but greater declines in planning performance, whereas implicit rewards yield weaker alignment with less degradation. This suggests that directly optimizing a style-specific objective can shift behavior at the expense of broader planning quality. Detailed settings and results are provided in Appendix~\ref{sec:supp_reward_ablation}.

\subsection{Safety Evaluation}
To assess safety after fine-tuning, we evaluate all strategy heads on 2,000 eight-second open-loop scenarios and observe low collision rates ranging from 0.85\% to 1.20\%, with small increases relative to the base planner (Table~\ref{tab:strategy_collision} in the Appendix).

\section{Conclusion}
We introduce M-Diffusion Planner, a novel interactive planning framework that leverages the diffusion model’s capacity to learn complex driving behaviors and generate diverse, high-quality trajectories. The framework exhibits strong flexibility and robustness across a wide range of driving scenarios. Building on the probabilistic nature of diffusion models, we incorporate GRPO to fine-tune a set of strategy modes. This allows the planner to adapt to various driving styles without significantly compromising overall driving performance. The planner supports trajectory generation conditioned on multi-modal user instructions, enabling behaviors aligned with user-specified driving modes. Crucially, it supports real-time strategy switching without requiring model retraining or reloading, offering both adaptability and efficiency for interactive autonomous driving applications. Open-loop and closed-loop experiments demonstrate that M-Diffusion Planner not only generates clearly differentiated strategy-specific trajectories, but also delivers competitive or superior planning performance compared with existing methods on the nuPlan Val14 benchmark.

\section{Limitations}
This study focuses on learning desired driving styles rather than defining the nature of driving style itself. Our formulation involves a key abstraction gap: driving style is implicitly represented through reward-weighted quantitative criteria, whereas real-world driving behavior is jointly shaped by experiential, normative, and situational factors. Consequently, representing driving style as a discrete label associated with a specific reward configuration oversimplifies a semantic and context-dependent behavioral space. With sufficient style-specific supervision, the framework could be extended beyond a limited set of canonical strategies to support richer, more personalized, and scenario-dependent driving styles while remaining compatible with continual training.

% Bibliography style is supplied by acl.sty (acl_natbib).
\clearpage
\bibliography{reference}

%%%%%%%%%%%%%%%%%%%%%%%%%%%%%%%%%%%%%%%%%%%%%%%%%%%%%%%%%%%%
\clearpage
\appendix
\raggedbottom

\section{Appendix}
\subsection{Diffusion sampling}
In this appendix, we detail the deterministic trajectory sampling procedure used during inference, as summarized in Algorithm~\ref{alg:inference_appendix}.
\begin{algorithm*}[t]
\caption{Deterministic Trajectory Sampling}
\label{alg:inference_appendix}
\begin{algorithmic}[1]

\State \textbf{Input:} Model $f_\theta$, initial observed state $x_0^{\text{obs}}$, context $c$, route $R$, mask $M$, strategy ID $s$

\State Construct initial sample:
\[
x_T \leftarrow \text{Concat}(x_0^{\text{obs}}, \epsilon), \quad \epsilon \sim \mathcal{N}(0, I)
\]

\State Reshape:
\[
x_T \in \mathbb{R}^{B \times P \times (T+1) \times d_s} \rightarrow \mathbb{R}^{B \times P \times d}
\]

\State Initialize solver with VP-SDE and score model $f_\theta$

\State Apply hard constraint to fix $x_0^{\text{obs}}$ during all steps

\For{$i = T$ \textbf{down to} $0$}

    \State \textbf{(1) Predict clean sample:}
    \[
    \hat{x}_0 \leftarrow f_\theta(x_t, t, c, R, M, s)
    \]

    \State \textbf{(2) Convert to score:}
    \[
    s_\theta(x_t,t) \leftarrow -\frac{x_t - \alpha(t)\,\hat{x}_0}{\sigma(t)^2 + \varepsilon}
    \]

    \State Update $x_{t_{i-1}}$ using a second-order multistep ODE solver with $s_\theta(x_t,t)$

    \State Enforce constraint:
    \[
    x_{t_{i-1}}^{(0)} \leftarrow x_0^{\text{obs}}
    \]

\EndFor

\State Reshape output:
\[
x_0 \rightarrow \mathbb{R}^{B \times P \times (T+1) \times d_s}
\]

\State Inverse normalize $x_0$, extract and return $\hat{x}_{1:T}$

\end{algorithmic}
\end{algorithm*}

\subsection{GRPO Training Procedure}
The overall reinforcement learning optimization procedure used in this work is summarized in Algorithm~\ref{alg:grpo_appendix}. Built upon the Group Relative Policy Optimization (GRPO) framework, the proposed training strategy optimizes the diffusion-based planning policy through trajectory-level reward supervision while maintaining policy stability via clipped policy updates and KL regularization.

\label{sec:appendix_grpo}

\begin{algorithm*}[t]
\caption{GRPO Training}
\label{alg:grpo_appendix}

\begin{algorithmic}[1]

\State \textbf{Input:} Data loader $\mathcal{D}$, diffusion policy $f_\theta$, reward function $R$, strategy ID $s$, sample count $G$, clipping parameter $\varepsilon$, KL weight $\beta$

\State Initialize frozen old policy:
$\pi_{\theta_{\mathrm{old}}} \leftarrow \pi_\theta$

\For{each batch $x$ in $\mathcal{D}$}

    \State Encode latent representation:
    $z \leftarrow \mathrm{Encoder}(x)$

    \State Sample $G$ trajectories from the frozen policy:
    $\hat{\tau}_1,\dots,\hat{\tau}_G
    \sim
    \pi_{\theta_{\mathrm{old}}}(z,s)$

    \For{$i=1$ to $G$}

        \State Compute trajectory reward:
        $r_i \leftarrow R(\hat{\tau}_i)$

    \EndFor

    \State Compute reward statistics:
    $\mu_r \leftarrow \mathrm{mean}(r_{1:G})$

    \State
    $\sigma_r \leftarrow \mathrm{std}(r_{1:G})$

    \For{$i=1$ to $G$}

        \State Compute normalized advantage:
        $
        A_i
        \leftarrow
        \frac{r_i-\mu_r}{\sigma_r+\epsilon}
        $

        \State Compute trajectory likelihood ratio:
        $
        \rho_i
        \leftarrow
        \frac{
        p_\theta(\hat{\tau}_i \mid z,s)
        }{
        p_{\theta_{\mathrm{old}}}(\hat{\tau}_i \mid z,s)
        }
        $

        \State Compute clipped GRPO objective:
        $
        \ell_i^{\mathrm{GRPO}}
        \leftarrow
        \min
        \left(
        \rho_i A_i,
        \mathrm{clip}(\rho_i,1-\varepsilon,1+\varepsilon)A_i
        \right)
        $

    \EndFor

    \State Compute policy objective:
    $
    \ell_p
    \leftarrow
    -
    \frac{1}{G}
    \sum_{i=1}^{G}
    \ell_i^{\mathrm{GRPO}}
    $

    \State Compute KL regularization:
    $
    \ell_{\mathrm{KL}}
    \leftarrow
    \mathbb{D}_{\mathrm{KL}}
    \left(
    \pi_\theta
    \parallel
    \pi_{\mathrm{ref}}
    \right)
    $

    \State Compute total loss:
    $
    \ell
    \leftarrow
    \ell_p
    +
    \beta \ell_{\mathrm{KL}}
    $

    \State Backpropagate gradients and update $\theta$

    \State Update EMA model

\EndFor

\State Synchronize old policy:
$\pi_{\theta_{\mathrm{old}}} \leftarrow \pi_\theta$

\State \textbf{Return:} Average training loss

\end{algorithmic}
\end{algorithm*}

\subsection{Detailed Framework Illustration}
\begin{figure*}[t]
    \centering
    \includegraphics[width=\textwidth]{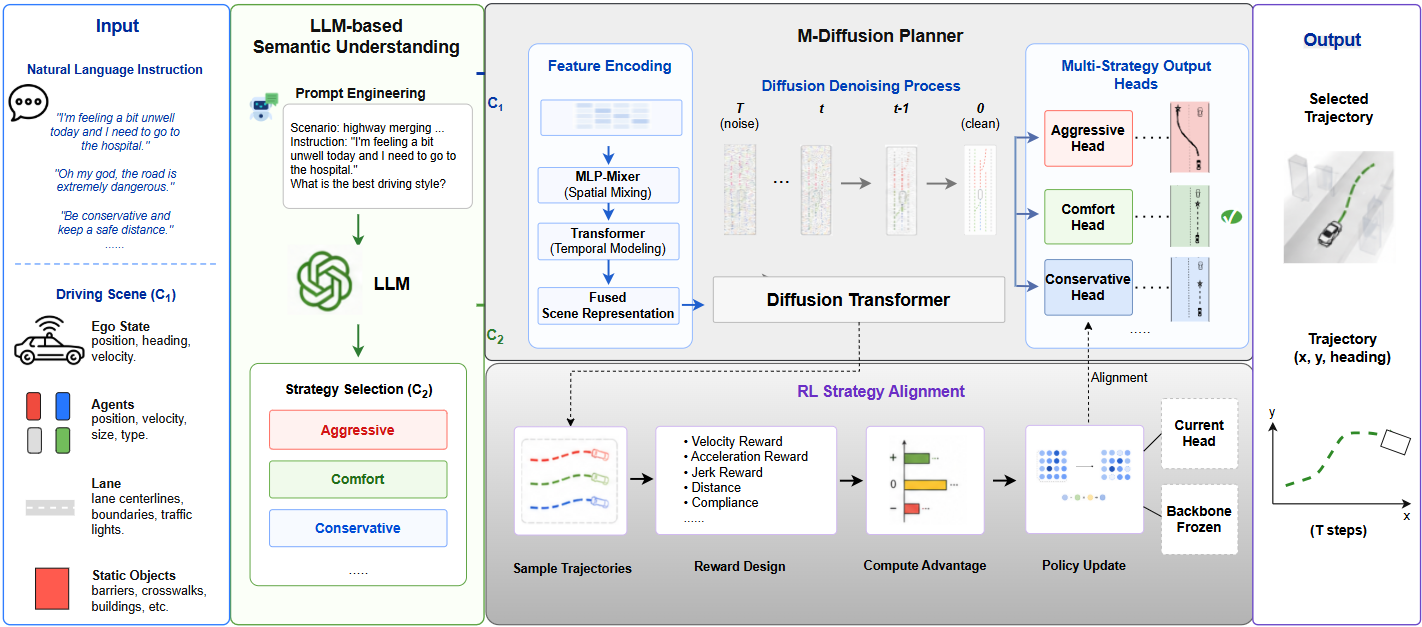}
    \caption{
    Overview of the proposed M-Diffusion framework with RL-based strategy alignment. 
    Given a driving scene and natural language instruction, an LLM first performs semantic understanding and selects the corresponding driving strategy. 
    The selected strategy activates the corresponding output head
    of the diffusion planner, which generates trajectories from
    the scene context using that head.
    To further align generated behaviors with different driving styles, a GRPO-based reinforcement learning optimization module is introduced, where sampled trajectories are evaluated using multiple reward functions and optimized through relative advantage estimation and policy updates. 
    The final planner produces trajectories that simultaneously satisfy scene constraints and high-level driving intentions.
    }
    \label{fig:framework}
\end{figure*}

Figure~\ref{fig:framework} illustrates the overall architecture and reinforcement learning optimization pipeline of the proposed framework. The framework integrates LLM-based intent understanding, multi-head diffusion planning, and GRPO-based strategy alignment. Specifically, the LLM maps a natural-language instruction to a discrete driving strategy $C_2$, which selects the corresponding output head $h(C_2)$ through a deterministic rule-based mapping. The diffusion planner takes the scene context $C_1$ as input and uses the selected head to predict clean trajectories during denoising. During GRPO post-training, only the selected strategy-specific output head is optimized using its corresponding reward function, while the shared backbone and other output heads remain frozen. This design provides separately optimizable strategy interfaces while sharing the pretrained planning backbone.

% Add \usepackage{graphicx} to the preamble.

\subsection{LLM-Based Intent Understanding and Strategy Selection}

\begin{figure}[t]
    \centering
    \includegraphics[width=\linewidth]{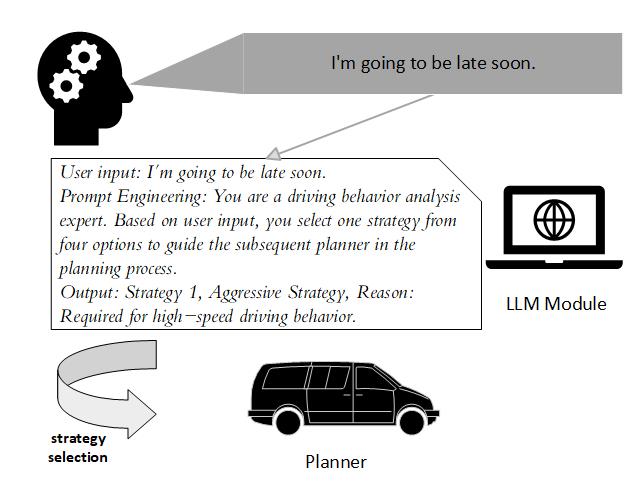}
    \caption{LLM-based intent understanding and strategy selection. The LLM interprets the user's natural-language input and selects a driving strategy, which activates the corresponding planner output head.}
    \label{fig:intent_strategy_selection}
\end{figure}

As illustrated in Figure~\ref{fig:intent_strategy_selection}, the LLM-based interaction module maps natural-language user input to an executable driving strategy. Given an instruction such as ``I'm going to be late soon,'' the LLM interprets the implied preference for travel efficiency and selects one of four predefined strategies using a structured prompt. In this example, it selects the aggressive strategy to prioritize efficiency within the planner's safety constraints. The selected strategy is then deterministically mapped to its corresponding output head, which generates trajectories aligned with that preference. When the user expresses a different need, the module updates the strategy selection and activates the corresponding head.

\subsection{Reward Design Ablation for Strategy Learning}
\label{sec:supp_reward_ablation}

In this section, we provide additional analysis on the impact of reward function design in the proposed multi-strategy learning framework. Specifically, we conduct an ablation study to investigate how different levels of reward specificity affect strategy formation, behavioral diversity, and overall performance.

\paragraph{Experimental Setup.}
We consider two categories of reward functions with different levels of explicitness:

\begin{itemize}
    \item \textbf{Explicit Rewards.}
    These rewards directly emphasize conservative driving
    objectives, including lower speeds and reduced acceleration
    magnitudes, encouraging clear behavioral alignment with
    the target strategy.

    \item \textbf{Implicit Rewards.}
    These rewards combine objectives for lower speeds and
    reduced acceleration magnitudes with broader goals such
    as safety, smoothness, and efficiency, without strongly
    enforcing a specific driving style.
\end{itemize}

All experiments are conducted under the same training configuration, with identical model architectures and datasets, to ensure a fair comparison.

\paragraph{Results and Analysis.}

Table~\ref{tab:reward_tradeoff_epoch} shows the trade-off between
conservative-strategy alignment and overall planning performance.
We evaluate average speed on 500 randomly sampled open-loop
scenarios and nuPlan scores on 100 closed-loop simulation scenarios.

\begin{itemize}
    \item \textbf{Explicit Reward Functions.}
    Explicit rewards produce a larger reduction in average speed,
    from 5.0160 to 4.0622 over 20 training epochs, indicating
    stronger speed-based alignment with the conservative strategy.
    However, the nuPlan score also decreases from 93.31 to 62.13,
    revealing substantial degradation in overall planning performance.

    \item \textbf{Implicit Reward Functions.}
    Implicit rewards induce a smaller reduction in average speed,
    from 5.0160 to 4.9386 over 20 training epochs, while the nuPlan
    score decreases from 93.31 to 78.61. Compared with explicit
    rewards, this configuration retains higher planning scores
    but achieves weaker speed-based alignment with the
    conservative strategy.
\end{itemize}

\begin{table}[t]
\centering
\small
\caption{Effect of explicit and implicit reward designs on
conservative-strategy alignment across training epochs.
Average speed is evaluated on 500 open-loop scenarios,
and nuPlan scores on 100 closed-loop scenarios.}
\label{tab:reward_tradeoff_epoch}
\resizebox{\linewidth}{!}{
\begin{tabular}{cccc}
\toprule
\textbf{Reward Design} &
\textbf{Epoch} &
\textbf{Avg. Speed} $\downarrow$ &
\textbf{nuPlan Score} $\uparrow$ \\
\midrule
\multirow{6}{*}{Explicit Reward}
& 0  & 5.0160 & 93.31 \\
& 3  & 4.6827 & 91.37 \\
& 5  & 4.4628 & 86.21 \\
& 10 & 4.2101 & 78.11 \\
& 15 & 4.1023 & 72.21 \\
& 20 & 4.0622 & 62.13 \\
\midrule
\multirow{6}{*}{Implicit Reward}
& 0  & 5.0160 & 93.31 \\
& 3  & 4.9621 & 92.67 \\
& 5  & 4.9613 & 88.21 \\
& 10 & 4.9526 & 83.91 \\
& 15 & 4.9447 & 80.21 \\
& 20 & 4.9386 & 78.61 \\
\bottomrule
\end{tabular}
}
\end{table}

\paragraph{Discussion.}
These findings highlight a fundamental trade-off in reward design: explicit rewards facilitate clear behavioral differentiation, but may introduce biases due to over-optimization toward specific objectives; meanwhile, implicit rewards do not significantly preserve the model’s foundational capabilities, and in the absence of strong behavioral alignment, the base capabilities also degrade simultaneously.

\paragraph{Implication.}
This study suggests that an effective strategy learning framework should balance these two aspects, potentially through hybrid reward design or adaptive weighting mechanisms, which we leave for future work.

%%%%%%%%%%%%%%%%%%%%%%%%%%%%%%%%%%%%%%%%%%%%%%%%%%%%%%%%%%%%
\subsection{Case Study}

This section provides additional details of the closed-loop experiments conducted in the \textit{nuPlan} simulator. In contrast to offline evaluation, the closed-loop setting continuously feeds the planned trajectory back into the environment, allowing future observations and interactions to depend on the model's previous decisions. Such a setup better reflects realistic autonomous driving deployment conditions and evaluates the long-term stability of the planning policy.

\begin{figure}[t]
    \centering
    \includegraphics[width=\linewidth]{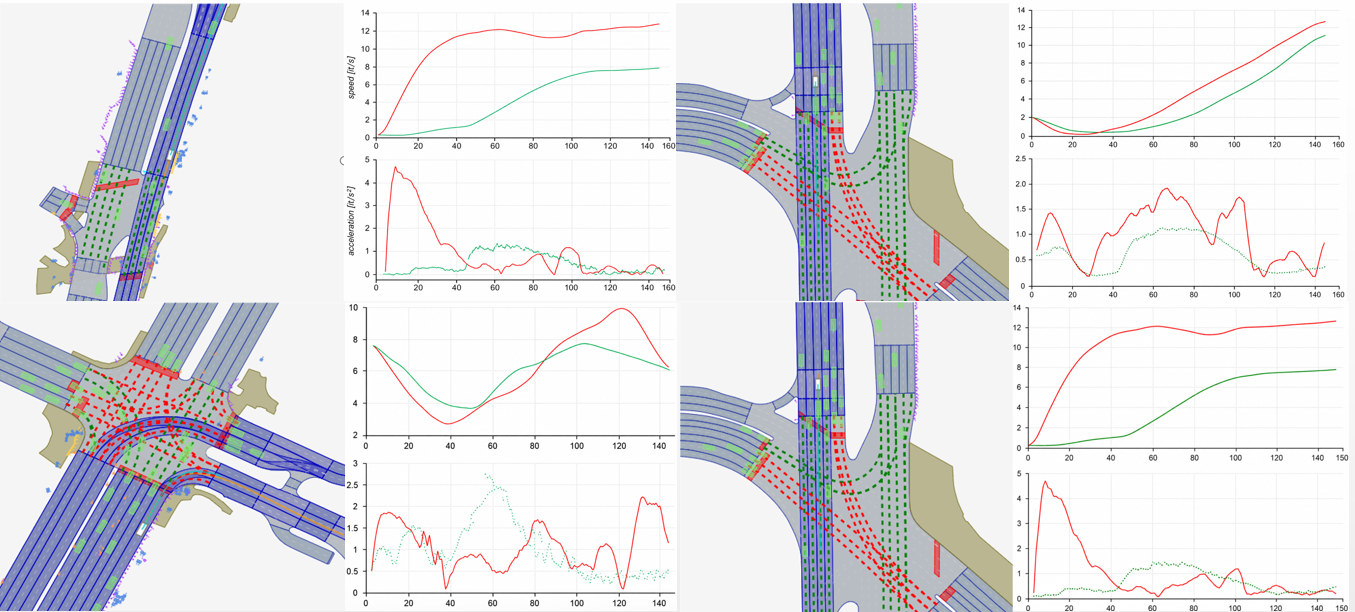}
    \caption{Representative closed-loop simulation cases in the \textit{nuPlan} environment under different driving strategies.}
    \label{fig:closed_loop_cases}
\end{figure}

Figure~\ref{fig:closed_loop_cases} illustrates representative closed-loop simulation results in the \textit{nuPlan} environment. Different driving strategies produce distinct interaction patterns and trajectory behaviors during long-horizon simulation, further demonstrating the controllability and adaptability of the proposed framework.

\subsection{More Details of Closed-Loop Experiments}

\begin{figure}[t]
    \centering
    \includegraphics[width=\linewidth]{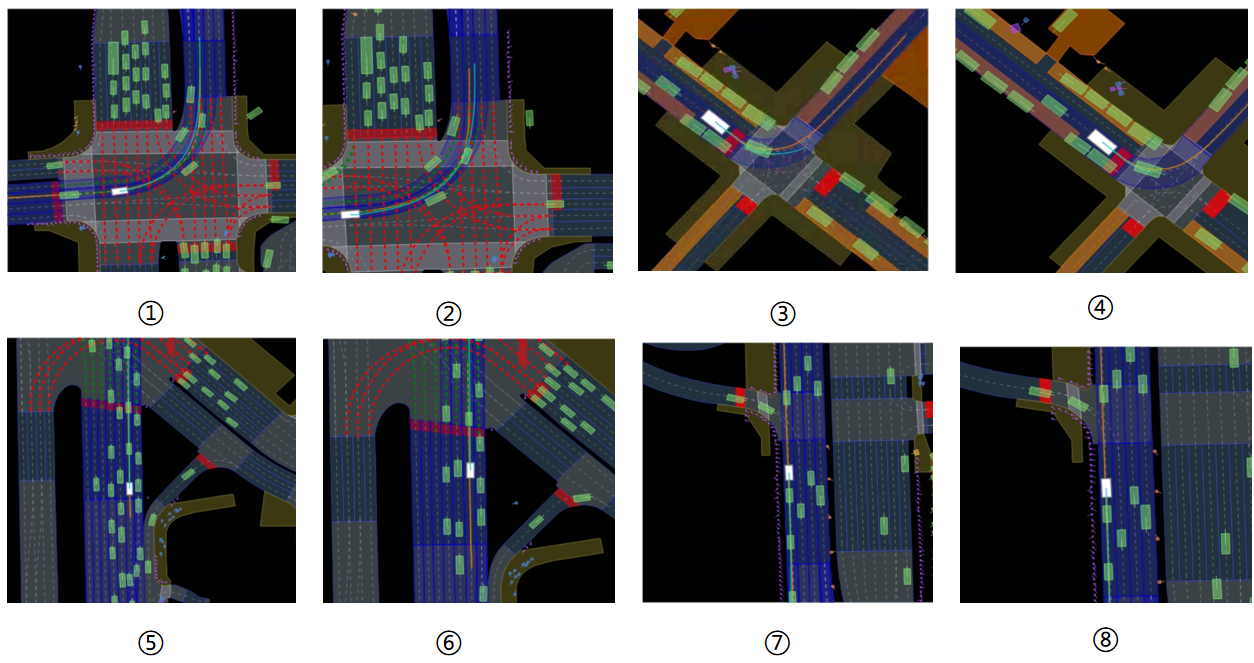}
    \caption{\centering Comparison between the base model and the conservative driving strategy under closed-loop simulation scenarios in NuPlan. Representative examples illustrate the behavioral differences in trajectory planning and motion patterns across different driving situations.}
    \label{fig:closed_loop_style_compare}
\end{figure}

To further analyze the behavioral differences under closed-loop simulation, we provide additional qualitative results of different driving styles generated by the proposed framework in the NuPlan simulator. As illustrated in Figure~\ref{fig:closed_loop_style_compare}, the outputs of the base planning model are compared with the conservative driving strategy under representative driving scenarios. The comparison highlights the differences in vehicle motion behaviors, including velocity adaptation and trajectory smoothness, demonstrating that the proposed strategy modulation mechanism can effectively influence the planning behavior while maintaining feasible closed-loop driving performance.

The simulation runs at 20 Hz, with the planner generating a new trajectory every 0.5 seconds. All models are evaluated under identical initial conditions and inputs. Scores range from 0 to 100, with higher values indicating better planning performance. All experiments are conducted on a workstation equipped with two NVIDIA RTX 4090 GPUs.

% \balance
\subsection{Additional Evaluations}
\label{sec:additional_evaluations}

\paragraph{Intent Understanding.}
We evaluate LLM-based strategy selection on 1,000 human-intent
pairs with ground-truth strategy labels, using the same prompt
and label space across models. As shown in
Table~\ref{tab:intent_accuracy}, all three models achieve
accuracy above 96\%.

\begin{table}[H]
\centering
\small
\caption{Intent classification accuracy on 1,000 human-intent pairs.}
\label{tab:intent_accuracy}
\begin{tabular}{lc}
\toprule
\textbf{Model} & \textbf{Accuracy (\%)} \\
\midrule
Gemma 3 12B  & 97.1 \\
GPT-4o      & 96.9 \\
Laguna S 2.1 & 97.6 \\
\bottomrule
\end{tabular}
\end{table}

\paragraph{Safety Evaluation.}
We assess open-loop collision rates on 2,000 randomly sampled
eight-second scenarios from the nuPlan test set.
Table~\ref{tab:strategy_collision} reports rates ranging from
0.85\% to 1.2\%, with all specialized strategies showing
slightly higher collision rates than the base planner.

\begin{table}[H]
\centering
\small
\caption{Open-loop collision rates on 2,000 eight-second scenarios.}
\label{tab:strategy_collision}
\begin{tabular}{lc}
\toprule
\textbf{Strategy} & \textbf{Collision Rate (\%)} \\
\midrule
Base         & 0.85 \\
Aggressive   & 1.20 \\
Conservative & 0.90 \\
Comfortable  & 1.10 \\
\bottomrule
\end{tabular}
\end{table}

\paragraph{REINFORCE-Based Post-Training.}
To investigate the effectiveness of conventional reinforcement learning for diffusion-based planning, we post-train the pretrained Diffusion Planner using REINFORCE with the same reward function as our GRPO-based approach. As shown in Table~\ref{tab:reinforce_results}, REINFORCE achieves a nuPlan score of 61.27 after three epochs. However, continued training leads to substantial performance degradation, with the score falling below 20 after ten epochs. These results suggest that directly applying conventional policy-gradient optimization may compromise pretrained planning capabilities, motivating our constrained GRPO-based strategy alignment.
\begin{table}[H]
\centering
\small
\caption{Planning performance during REINFORCE-based alignment.}
\label{tab:reinforce_results}
\begin{tabular}{lcc}
\toprule
\textbf{Method} & \textbf{Epochs} & \textbf{nuPlan Score} \\
\midrule
REINFORCE & 3  & 61.27 \\
REINFORCE & 10 & $<20$ \\
\bottomrule
\end{tabular}
\end{table}

\end{document}